%% file: main.tex
\documentclass[defaultstyle,11pt]{article}

\usepackage{amssymb}		
\usepackage{graphicx}		
\usepackage{hyperref}		

\title{A Random Point Initialization Approach to Image Segmentation with Variational Level-sets}

\def\jem{Postal Address:
      Department of Applied Mathematics,
      University of Colorado, Box 526
      Boulder CO 80309-0526, USA; email: corcoran@colorado.edu, 
      phone: 303-492-0685}

\author {J.N. Mueller, J.N. Corcoran\\
University of Colorado Boulder \thanks{\jem} }

\input macros.tex

\begin{document}

\maketitle
\vspace{-.8cm}

\begin{abstract} \small \noindent

Image segmentation is an essential component in many image processing and computer vision tasks. The primary goal of image segmentation is to simplify an image for easier analysis, and there are two broad approaches for achieving this: edge based methods, which extract the boundaries of specific known objects, and region based methods, which partition the image into regions that are statistically homogeneous. One of the more prominent edge finding methods, known as the \emph{level set method}, evolves a zero-level contour in the image plane with gradient descent until the contour has converged to the object boundaries. While the classical level set method and its variants have proved successful in segmenting real images, they are susceptible to becoming stuck in noisy regions of the image plane without \emph{a priori} knowledge of the image and they are unable to provide details beyond object outer boundary locations. We propose a modification to the variational level set image segmentation method that can quickly detect object boundaries by making use of \emph{random point initialization}. We demonstrate the efficacy of our approach by comparing the performance of our method on real images to that of the prominent Canny Method.

\bigskip
\end{abstract}

\setcounter{page}{0}

\section{Introduction}
\label{s:intro}

Image segmentation is one of the most challenging tasks in image processing, which is essential for myriad computer vision tasks including facial recognition, traffic control systems management, autonomous vehicle development, robotics, and medical imaging and tomography. The primary goal of image segmentation is to simplify the image for easier analysis. This can be done in several ways, for example, by breaking up an image into homogeneous regions made up of pixels with some measure of similarity or detecting edges based on continuity changes in pixel intensity.

The development of the first techniques for image segmentation can be traced back more than 50 years. In 1965, an operator for detecting edges between different parts of an image, known as the the {\it{Roberts edge detector}} (also referred to as the Roberts operator), was introduced \cite{roberts1965} as a first step toward decomposing an image into its constituent components. Since the introduction of this operator, many approaches to image segmentation that range from clustering algorithms to curve evolution have been discovered. Broadly speaking, they can be split into two categories. The first is to find the boundaries of specific known objects, for example, in MRI scans. Approaches on this front have been very successful as in \cite{bouman94}, \cite{cheng2016}, and  \cite{puonti}. The second problem is to partition an image into disjoint regions that are statistically homogeneous. Based on an image alone,  this problem is generally ill-posed with potentially multiple reasonable partitions. This is especially true for low quality images where a lack of assumptions about smoothness will result in a noisy segmentation. 

\section{Active Contours}
\label{s:lsm} 
The active contour approach to edge detection is a partial differential equation (PDE)-based segmentation method that treats image segmentation as an energy optimization task. Broadly speaking, an initial curve (or set of curves) is defined and driven into the interior of the image plane along the negative gradient of an energy functional using gradient descent methods  \cite{chan2001, li2005, li2010, weeratunga2003, yuan2012}. In particular, the energy functional is composed of terms representing the \emph{internal energy} of the evolving contour and \emph{external energy} of the image \cite{chan2001}. The external energy is computed from underlying image properties such as image gradients and acts as a ``shrinkage" term, drawing the contour into the image plane toward object boundaries, while the internal energy acts as a regularization term which controls the smoothness and continuity of the deforming curve \cite{chan2001, yuan2012, cremers2007}. The energy functional must be carefully chosen so that the propagating curve stops once it encounters object boundaries within the image. Often the chosen active contour models include a gradient dependent edge-function, $g$, since object boundaries in the image occur at the minimum (or critical) value(s) of the energy functional \cite{chan2001}.  

Commonly, active contour methods either construct the initial curve explicitly as a parameterized set of Lagrangian curves or implicitly represent the initial curve as a certain contour of a higher dimensional function in an Eulerian framework \cite{li2005}. Explicit active contour methods were first introduced in \cite{kass1987} and are robust with respect to image noise and boundary gaps \cite{weeratunga2003}. However, explicit parameterization schemes are often sensitive to the initial conditions used and struggle to handle significant changes in curve geometry such as merging or splitting that may naturally arise in an image segmentation task. Thus, Lagrangian methods rely on a continual reparameterization, which usually becomes more complex as the segmentation task progresses, and they can break down as the evolving curve's shape changes.  Methods which handle such topological changes naturally and avoid the complexities of reparameterization alternately define the curve implicitly as a level set.

\subsection{Level Set Methods}
The \emph{level set method} was introduced in \cite{sethian1977} as a means of following complicated dynamics in front propagation and was separately extended to the task of edge detection in \cite{caselles1993} and \cite{malladi1995}. The basic idea is to represent the initial curve implicitly by embedding it as the \emph{zero-level contour} (or zero-level set) of a higher-dimensional hypersurface known as the \emph{level set function} and denoted $\varphi$. That is, 
we define the curve of interest to be the contour
$$
    \mathcal{C} := \big\{ \vec{x} \ | \ \varphi(\vec{x},t) = 0 \big\}.
$$
The level set function (and, thus, the zero-level set curve) is optimized according to a level set evolution equation such as
\begin{align}
    \frac{\partial \varphi}{\partial t} 
        &= - \frac{\partial E}{\partial t}
         = - F \vert \nabla \varphi \vert \, .
\label{e:lsm_evolequ}
\end{align}

A detraction of traditional level set methods is the shape of the curve can become distorted as it evolves in the image plane so that portions of the curve form ``shocks," becoming excessively steep or flat over time \cite{li2005, suri2002}. The formation of shocks can result in instability in the algorithm if unchecked and the computed solution where shocks form is inaccurate. It follows the evolving curve must be periodically reshaped (known as ``re-initialization") according to some metric describing how deformed the curve has become \cite{li2010}. However, there are no systematic criteria for determining when or how often a curve should be reshaped \cite{Gomes2000} and in some cases reshaping can introduce additional error by spatially shifting the reformed curve away from the true hypersurface \cite{li2005}.

While early level set methods are ``pure PDE methods" which construct the evolution equation \eqref{e:lsm_evolequ} by first defining an evolution PDE in the Lagrangian framework and then converting it to an evolution PDE for the level set function \cite{li2005}, a method known as \emph{variational level sets} obtains the desired evolution equation by minimizing an energy functional defined on the level set function itself \cite{li2005, yuan2012, cremers2007}. First proposed in \cite{mumford1989}, this approach was popularized by \cite{chan2001}. Not only does the variational level set approach allow the evolution equation to be obtained in a more straightforward manner, but it also allows additional details such as region-based information \cite{chan2001, zhao1996} or shape information \cite{vemuri2003} to be included in the energy functional.

More recently, variational level set methods such as \cite{li2005}, \cite{li2010}, \cite{Gomes2000}, and \cite{liu2011} have been used to address the problem of shock formation. In particular, these methods which have emerged avoid the creation of shocks by including \emph{distance regularization terms} in the energy functional. The purpose of the regularization terms is to locally monitor the evolving curve and to penalize deviations away from a \emph{signed distance function} which may eventually lead to shock formation.\footnote{The choice of penalty term reflects the usual method of initializing the level set curve $\varphi_{0}$ as a signed distance function.}

For example, the variational form introduced in \cite{li2005} and expanded upon in \cite{li2010} exploits the property that any signed distance function satisfies $\vert \nabla \varphi \vert = 1$ to define a metric
\begin{align}
    \mathcal{P}
        &= \int_{\Omega} \frac{1}{2} \big( \vert \varphi \vert - 1)^{2} d\vec{x}
\end{align}
which measures deviations of the level set curve $\varphi$ from a signed distance function in a neighborhood $\Omega \subset \mathbb{R}^{2}$ around the desired zero-level set. This term depends only on $\varphi$ and represents the internal energy of the curve.

The energy functional which must be minimized is then given by a \emph{variational formula} 
\begin{align}
    E &= \mu \mathcal{P} + \mathcal{E}_{\text{\tiny ext}} 
    \label{e:lsm_energy}
\end{align}
where $\mu > 0$ controls the effect of the penalty on the evolving curve and $\mathcal{E}_{\text{\tiny ext}}$ is comprised of external and boundary energies that depend only on the image data. 

The gradient flow which propagates the level set curve $\mathcal{C}$ into the image plane is computed using a first G$\hat{\text{a}}$teaux derivative \cite{li2005, li2010} of the energy functional \eqref{e:lsm_energy}
\begin{align}
    \frac{\partial \varphi}{\partial t} 
        &= - \frac{\partial E}{\partial \varphi}
        = -\mu \frac{\partial \mathcal{P}}{\partial \varphi} - \frac{\partial \mathcal{E}_{\text{\tiny ext}}}{\partial \varphi} \, .
    \label{e:pikachu}
\end{align}
Specific expressions for $\mathcal{E}_{\text{\tiny ext}}$ and the completely differentiated right-hand size of \eqref{e:pikachu} are given in \cite{li2005} while finite difference schemes for implementing this method can be found in \cite{li_code, pikachu_code}.

\section{The Random Point Initialization Method}
\label{s:intro_rpi}

\subsection{Motivation}
\label{s:motive_rpi}

\begin{figure}[!h]
    \centering
    \includegraphics[width=\textwidth]{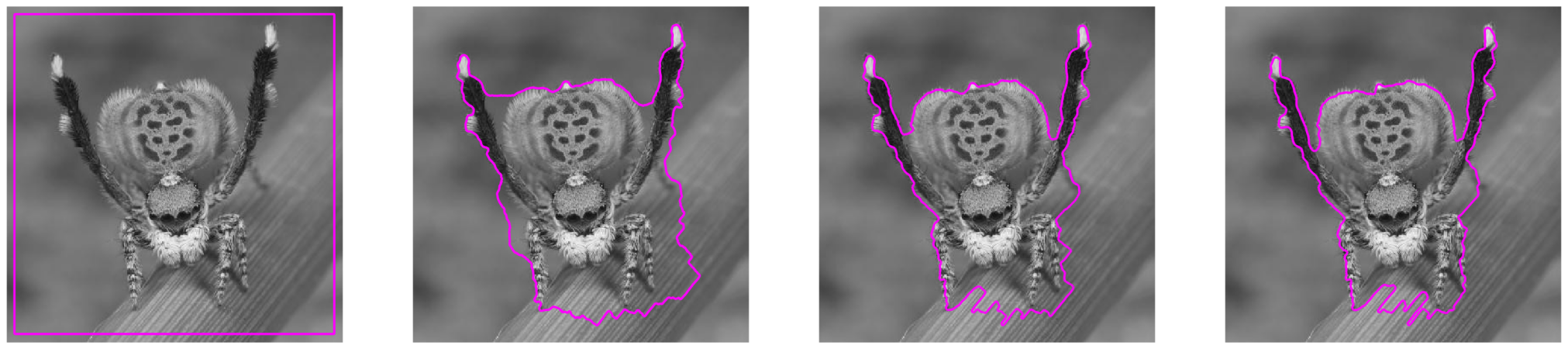}
    \caption{\emph{\bf Left} Gray-scale image of spider with initial level set curve (pink). \emph{\bf Center-left} Level set curve (pink) after 500 steps. \emph{\bf Center-right} Level set curve (pink) after 1,000 steps. \emph{\bf Right} Level set curve (pink) function after 50,000 steps.}
    \label{fig:spider_lsf}
\end{figure}

We motivate our modification to the variational level set method described in \cite{li2005} with the following example. Consider the image of a spider \cite{tsankashvili2018} given in the far-left frame of Figure \ref{fig:spider_lsf}. To segment the image and obtain the boundary of the spider, define an initial level set curve $\varphi_{0}$ as a step function
\begin{align}
    \varphi_{0} 
        &= \begin{cases} \ \ c_{0}, \quad \text{outside contour} \\ 
                        -c_{0}, \quad \text{inside contour} \end{cases}
\end{align}
where $c_{0} > 0$ is constant and drive the curve into the image plane along the energy gradient given by \eqref{e:pikachu} as in \cite{li_code, pikachu_code}.

Although the level set appears to converge quickly to the object boundary where the foreground (spider) and background are reasonably distinct and the image gradient is smooth,
the evolving curve stagnates around the spider's feet; even after tens of thousands of iterations the position of the level set is relatively unchanged.

Recognizing that the curve becomes stuck in regions where the image gradient is not smooth, we attempted to prompt convergence of the curve to the object boundary in these regions by periodically ``kicking" the points of the level set by a small, random amount $\boldsymbol{\varphi_{k}} \mapsto \boldsymbol{\varphi}_{k}+\varepsilon, \ \ \varepsilon \sim N(0,\sigma^{2})$. However, perturbing the level set in this way did not result in a significant improvement to the convergence speed. Perturbations large enough to noticeably displace the level set destroyed the curve and, while future iterations collected the scattered points into a salient boundary once again, experimental evidence suggests that the reformed boundary remained stuck in troublesome regions of the image. On the other hand, perturbations which locally displaced points were usually too small to nudge the evolving curve to regions in the image where the gradient were smooth.

\begin{figure}[h]
    \centering 
    \includegraphics[width=\textwidth]{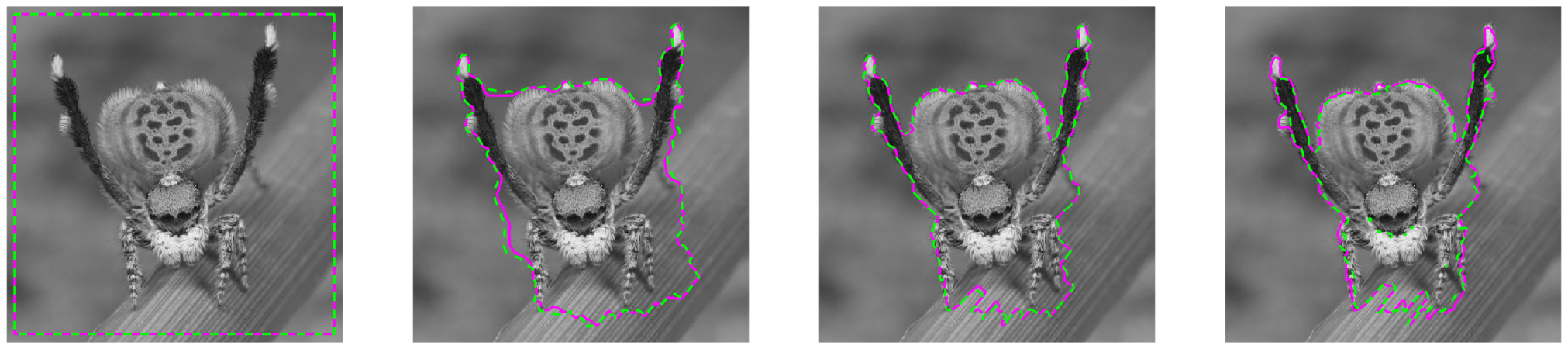}
    \caption{\emph{\bf Left} Gray-scale image of spider with initial level set curves (pink solid: iterated with variational level set method; green dashed: iterated with perturbed level set method). \emph{\bf Center-left} Level set curves after 500 steps. \emph{\bf Center-right} Level set curves after 1,000 steps. \emph{\bf Right} Level set curves after 50,000 steps.}
    \label{fig:spider_lsf_jit}
\end{figure}
 
Additionally, the segmentation obtained by periodically perturbing the level set often suffered from boundary leakage since it is possible for the perturbation to push points on the evolving curve across the object boundary. This is illustrated in the far-right frame of Figure \ref{fig:spider_lsf_jit}, wherein the white tops of the spider's feet are excluded from the segmentation obtained with a perturbed level set (green dashed) curve.
 
\subsection{The Random Point Initialization}
\label{s:rpi_method}
 
Our efforts to perturb the level set in this way provided the following key insight: \emph{points scattered by random perturbations are collected together in future iterations of the algorithm.} 
Edge detection algorithms depend on the image gradients to capture features of the underlying image topography and use this information to drive the defined hypersurface and the embedded level set curve to object boundaries (i.e. global minima) within the image plane. Importantly, the algorithm uses this gradient information to propagate any collection of points, regardless of whether these points define a coherent curve or are randomly scattered. 
If the algorithm iterates a curve, then the net affect is to contract the curve around the object(s) of interest (i.e. a lassoing effect). If instead the algorithm iterates a set of free points, the net affect is to push the points ``downhill" along the image gradients until they settle into minima of the image plane.

 \begin{figure}[!h]
\centering 
    \includegraphics[width=0.375\textwidth]{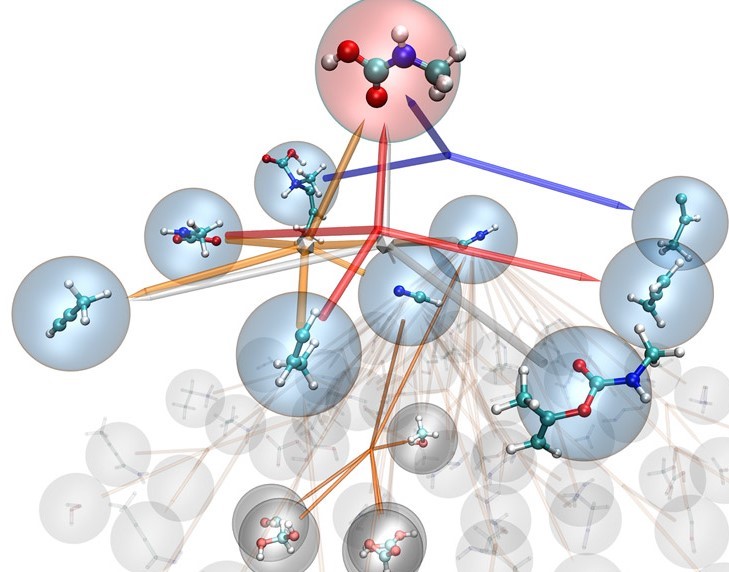}
\caption{A nano-system of chemical reactions.}
\label{fig:nanochem}
\end{figure}

We take advantage of this and make the following modification to the algorithm: instead of embedding a level set in a hypersurface as in \eqref{e:pikachu}, embed an initial set of randomly scattered points $\boldsymbol{\phi} \in \mathbb{R}^{I \times J}$ where $\phi_{ij} \sim N(0, \sigma^{2})$ and $\sigma$ is the size of the perturbation. The desired object boundary (edge-set) resulting from iterating $\boldsymbol{\phi}$ with gradient descent is still given by the zero-level set contour after $k$ steps.  We call this the \emph{random point initialization} (RPI) method.

\begin{figure}[!h]
\centering
    \includegraphics[width=\textwidth]{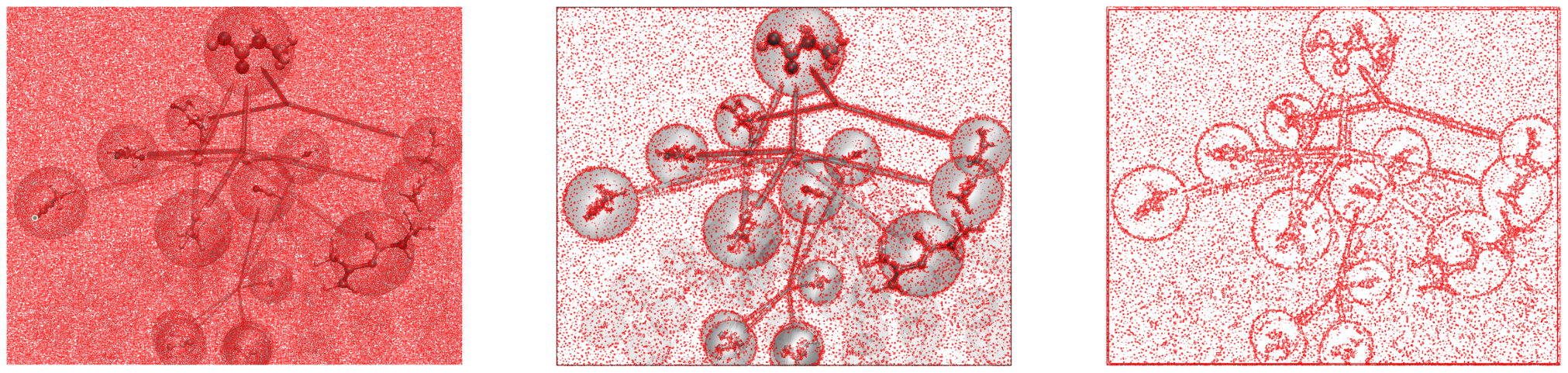}
\caption{\emph{Left:} Random points (red) scattered on top of the grayscale image. \emph{Center:} The result of $k=4$ steps of the variational level set method starting from randomly placed points, plotted with the original grayscale image for reference. \emph{Right:} The result of $k=4$ steps of the variational level set method starting from randomly placed points. The grayscale image has been removed from the background to make the coalescence of boundary points more apparent.}
\label{fig:chem_rpi}
\end{figure}

To illustrate the RPI approach clearly, consider the image of a system of chemical reactions depicted in Figure \ref{fig:nanochem} \cite{wang2014}. Scattering points randomly in the image plane (viewed on top of the grayscale image for reference) and completing several gradient descent steps using \cite{li_code, pikachu_code} results in the noisy segmentation shown in the far-right frame of Figure \ref{fig:chem_rpi}. 
Clearly, the points coalesce about the desired object boundaries. However, the segmented image is excessively noisy and the obtained boundary is diffuse. The problem of filling boundary gaps is easily solved by constructing the desired edge set from multiple runs of the RPI method. We demonstrate this idea first by plotting multiple edge sets obtained with the RPI method on top of one another. (The background image is omitted for clarity.) 
\begin{figure}[!h]
\centering
    \includegraphics[width=\textwidth]{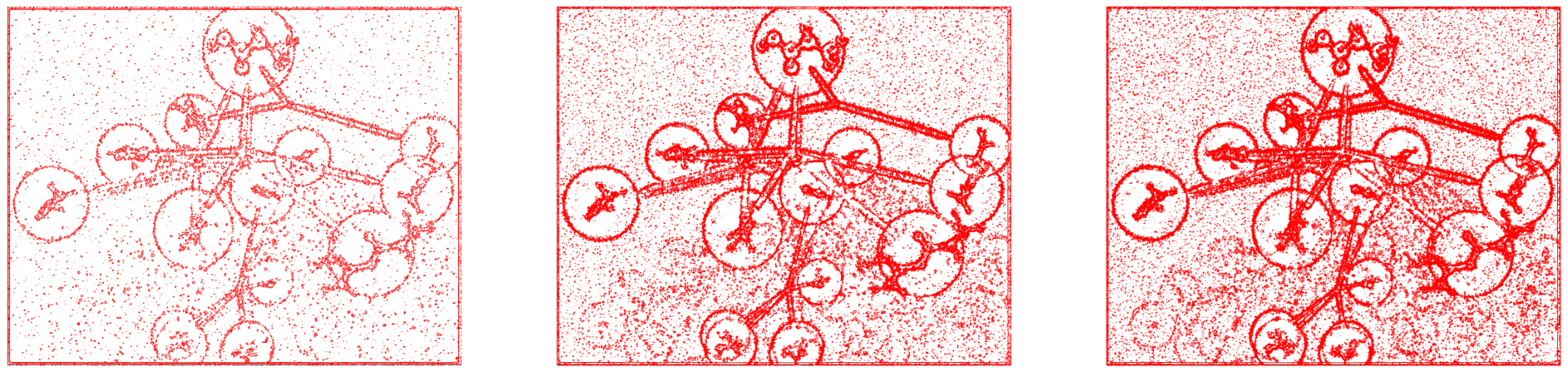}
\caption{Plotting the edge sets resulting from multiple RPI runs on top of one another. \emph{Left:} A single RPI zero-level set. \emph{Center:} Four RPI zero-level sets plotted on top of one another. \emph{Right:} Eight RPI zero-level sets plotted on top of one another.}
\end{figure}
Notice that as the zero-level sets are layered together the boundaries around objects within the image become more distinct. 

To replicate this outcome in a systematic way, we periodically (after $k$ iterations) re-initialize the evolving set as a random matrix $\boldsymbol{\phi}_{0}$. Each time we re-initialized the set to $\boldsymbol{\phi}_{0}$ and perform $k$ iterations, we complete a single run of the RPI method. The desired edge set from each run is the zero-level set, as before, which can be flattened into an associated vector of values.  

\begin{align*}
\boldsymbol{\phi}_{k}  =
\begin{bmatrix}
    \phi_{11}   & \phi_{12} & \hdots & \phi_{1J} \\
    \phi_{21}   & \phi_{22} & \hdots & \phi_{2J} \\
    \vdots & \vdots & \ddots & \vdots \\
    \phi_{I1} & \phi_{I2} & \hdots & \phi_{IJ}
\end{bmatrix}
\longrightarrow 
\begin{bmatrix}
    \phi_{11} \hdots \phi_{1J} &
    \phi_{21}\hdots\phi_{2J} &
    \hdots &
    \phi_{I1} \hdots \phi_{IJ}
\end{bmatrix}    
\end{align*}

We collect the flatten the edge sets from each RPI run by stacking the vectors vertically in a segmentation matrix. Here the super-script denotes the RPI run.

\begin{align*}
\begin{bmatrix}
    \boldsymbol{\phi}_{k} ^{(1)} \\
    \boldsymbol{\phi}_{k} ^{(2)} \\ 
    \vdots \\ 
    \boldsymbol{\phi}_{k} ^{(m)}
\end{bmatrix}
&=
\begin{bmatrix}
    \phi_{11}^{(1)} \hdots \phi_{1J}^{(1)} & \hdots & \phi_{I1}^{(1)} \hdots \phi_{IJ}^{(1)} \\
    \phi_{11}^{(2)} \hdots \phi_{1J}^{(2)} & \hdots & \phi_{I1}^{(2)} \hdots \phi_{IJ}^{(2)} \\
        \vdots  &\ddots & \vdots \\ 
    \phi_{11}^{(m)} \hdots \phi_{1J}^{(m)} & \hdots & \phi_{I1}^{(m)} \hdots \phi_{IJ}^{(m)} \\
\end{bmatrix}     \\
&\hspace{0.35in}
\begin{matrix}
    \downarrow & \textcolor{white}{\hdots} & \downarrow & \textcolor{white}{\hdots} & \textcolor{white}{\hdots} & \downarrow & \textcolor{white}{\hdots} & \downarrow
\end{matrix} \\
\overline{\boldsymbol{\phi} } 
&= 
\hspace{0.1in}
\begin{bmatrix}
    \overline{\phi}_{11} \hspace{0.025in} \hdots \hspace{0.025in} \overline{\phi}_{1J} \hspace{0.025in} & \hdots & \hspace{0.1in} \overline{\phi}_{I1} \hspace{0.05in} \hdots \hspace{0.05in} \overline{\phi}_{IJ}
\end{bmatrix}
\end{align*}

Once the iterations have been collected we perform element-wise averaging of the sets (that is, the average of each column) to find the mean ``image pixel" value in the segmentation matrix. The vector that results from this averaging process can then be reshaped into an $I \times J$ segmentation matrix that is the zero-level set of interest.
 
\begin{align*}
\overline{\boldsymbol{\phi} } =
\begin{bmatrix}
    \overline{\phi}_{11} \hspace{0.025in} \hdots \hspace{0.025in} \overline{\phi}_{1J} \hspace{0.025in} & \hdots & \hspace{0.1in} \overline{\phi}_{I1} \hspace{0.05in} \hdots \hspace{0.05in} \overline{\phi}_{IJ}
\end{bmatrix}
\longrightarrow
\begin{bmatrix}
    \overline{\phi}_{11}   & \overline{\phi}_{12} & \hdots & \overline{\phi}_{1J} \\
    \overline{\phi}_{21}   & \overline{\phi}_{22} & \hdots & \overline{\phi}_{2J} \\
    \vdots & \vdots & \ddots & \vdots \\
    \overline{\phi}_{I1} & \overline{\phi}_{m2} & \hdots & \overline{\phi}_{IJ}
\end{bmatrix}
\end{align*} 

\begin{figure}[!h]
\centering 
    \includegraphics[width=0.375\textwidth]{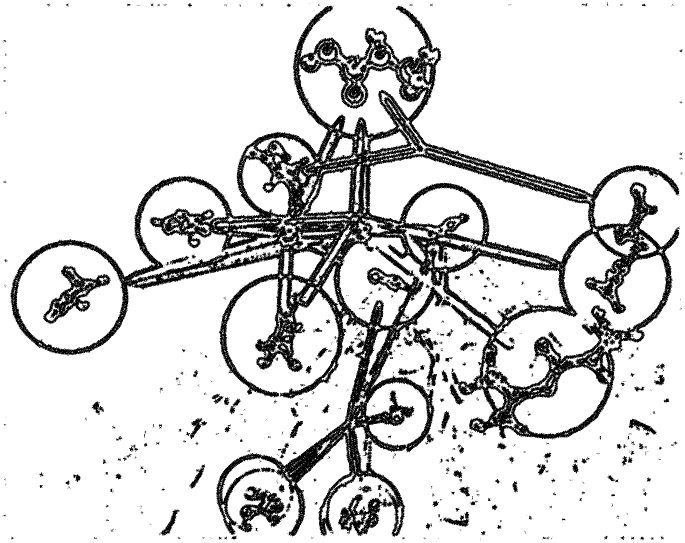}
\caption{Segmentation of the chemical system with $m$-random point initialization steps.}
\label{fig:nanochem_rpi}
\end{figure}

The segmentation of the chemical system resulting from combining multiple runs of the RPI method (including several post-processing steps discussed in Sec. \ref{ss:mrpi_postproc}) is given in Figure \ref{fig:nanochem_rpi}.

We emphasize that the random re-initialization is \emph{not} the same as the re-initialization methods found in the literature and discusses in Section \ref{s:lsm}, but is instead a numerically expedient means of repeatedly segmenting the same image and combining the resulting edge sets.

\subsection{Edge set Post Processing}
\label{ss:mrpi_postproc}

Once the pixel-averaged segmentation matrix $\boldsymbol{\overline{\phi}}$ has been constructed from multiple RPI runs, we found that the edge set usually contains some amount of noise from random points settling in local minima of the image and becoming stuck. Furthermore, we noticed that some points generally settled very close to the true boundary (true zero-level set) but were not pushed all the way into the minimum by the $k$th iteration. Further de-noising and smoothing of the segmented image is done with a series of post-processing steps.  

\subsubsection{Collecting Near-Boundary Points}

We collect near-boundary points and include them in the edge set by thresholding, i.e. defining a small range of values around the zero-level contour that we wish to include in the edge set. The set of points found in this range, which we denote $\boldsymbol{\overline{\phi}_{\text{\tiny Th}}}$, are then used as an approximation to the true zero-level set. 

As an aside, we found it was most effective to first normalize the values in $\boldsymbol{\overline{\phi}}$ before extracting the (approximate) zero-level set $\boldsymbol{\overline{\phi}_{\text{\tiny Th}}}$. We used the normalization
\begin{align}
\dfrac{\boldsymbol{\overline{\phi}}}{||\boldsymbol{\overline{\phi}}||} 
    &= \dfrac{\boldsymbol{\overline{\phi}}}{\max_{i,j}|\boldsymbol{\overline{\phi}}| + \min_{i,j}|\boldsymbol{\overline{\phi}}|}
    \label{e:norm_phi}
\end{align}
but we note that other choices for normalization, such as dividing by the absolute maximum pixel value, appeared to yield similarly good results.

Note that choosing
\begin{align}
\boldsymbol{\overline{\phi}_{\text{\tiny Th}}} 
    = \bigg\{ \frac{\boldsymbol{\overline{\phi}}}{||\boldsymbol{\overline{\phi}}||} \ \bigg \vert \  p_{\text{\tiny low}} \leq \frac{\boldsymbol{\overline{\phi}}}{||\boldsymbol{\overline{\phi}}||}  \leq p_{\text{\tiny up}} \bigg\}
\label{e:phi_log}
\end{align}
eliminates much of the noise in the edge set, as points which have settled in local minimum values will be automatically excluded from this set.

Obtaining the edge set in \eqref{e:phi_log} using conditional relationships results in an edge set of binary entries of the form
$$
    (\overline{\phi}_{i,j})_{\text{\tiny Th}} 
        = \begin{cases}
            1, \qquad \overline{\phi}_{i,j} \in [p_{\text{\tiny low}}, p_{\text{\tiny up}}] \\
            0, \qquad \text{otherwise}
        \end{cases}
$$
where pixels of value $1$ are black and pixels of value $0$ are white. The final post-processing steps used to thin and smooth the edge set require binary entries such as these.

\subsubsection{Further Curve Smoothing and Image De-noising}
\label{s:major_thin}

Segmentation algorithms are evaluated by both the accuracy 
and quality of the obtained boundary, usually defined by the thickness and continuity of the curve in the edge set. Since the multiple-RPI method is based on collecting free points along edges within the image plane, the obtained segmented image often has an undesirably thick boundary. Once the zero-level edge set $\ds \boldsymbol{\overline{\phi}_{\text{\tiny Th}}}$ is constructed we further refine the segmented image by smoothing and thinning the edges. 

The edges are smoothed using morphological operations which examine small non-overlapping 
neighborhoods of pixels. In particular, the operator compares the central pixel $p$ in a $3 \times 3$ neighborhood to its eight-neighbor pixels and assigns to $p$ the value ($0$ or $1$) shared by a majority of its neighbors. In the event of a tie, $p$ is treated as a pixel of noise and is assigned $p=0$ \cite{thompson1995image}. This removes extraneous pixels from the segmented image while also filling in small spurious gaps in the boundary.

The edges can be thinned with a ``skeletonizing" algorithm such as \cite{lam1992} (first introduced in \cite{zhang1984} but later modified in \cite{lam1992} to reduce the likelihood of introducing boundary gaps). 
The thinning algorithm uses contour-following methods to count the number of times a pixel is traced. Pixels only traced once are deleted (set to $0$), while pixels traced multiple times are considered essential to maintaining connectivity in the boundary (keep the value of $1$). In a basic sense, the algorithm thins the boundary while retaining two fundamental types of geometry: diagonal lines and $2 \times 2$ squares \cite{lam1992, zhang1984}. 
While the algorithm may be run until the image stops changing, in practice it is often only necessary to run two or three iterations to sufficiently reduce the thickness of the boundary found in these post-processing steps.

\subsection{Other modifications}
\label{ss:othermod}
At times scatting $IJ$ random points in the image plane can lead to points piling up in certain locations of the image plane, resulting in object boundaries that are excessively thick or that merge with the boundaries of nearby objects. When this happens, it may obscure subtle details in the image that we may wish to capture. 

One solution which often improves the result of segmenting with the $m$-RPI approach is to re-size the image using bicubic interpolation to spatially separate objects within the image. By pushing the boundaries of objects apart from one another, points scattered in the image plane collect around individual objects since the objects are distinct and reduces the likelihood of merged boundaries.

Alternately, the number of random points scattered in the image plane may be reduced by some fraction $\alpha$ to $\alpha IJ, \ \ 0 < \alpha < 1$, of the total number of pixels in the image.
\begin{figure}[!h]
    \centering
    \begin{minipage}{0.3\textwidth}
        \includegraphics[width=0.75\textwidth, height=0.18\textheight]{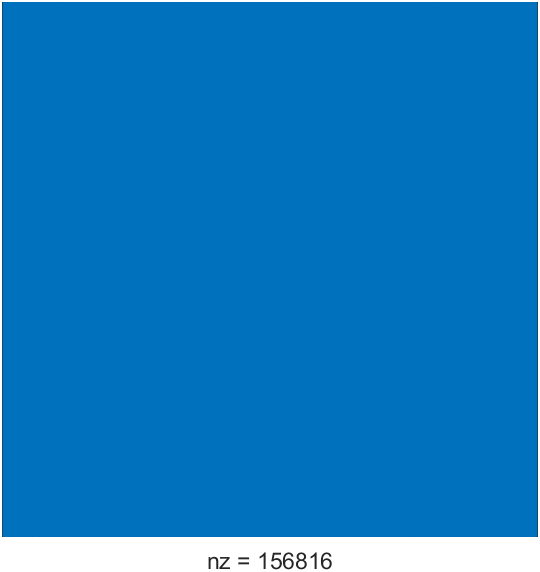}
    \end{minipage}%
$\longrightarrow$     
    \begin{minipage}{0.3\textwidth}
        \includegraphics[width=0.75\textwidth, height=0.18\textheight]{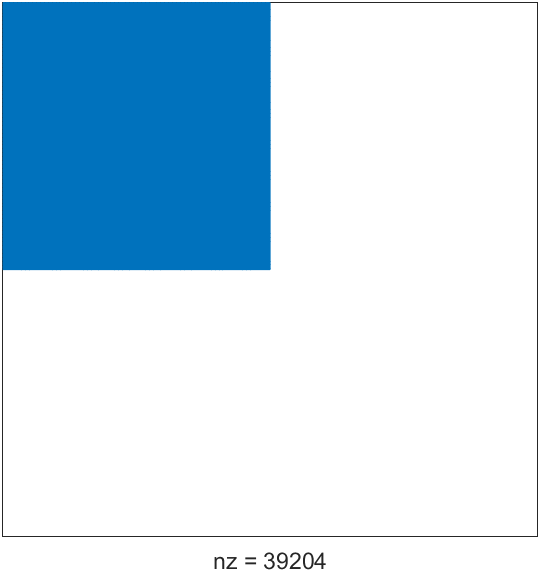}
    \end{minipage}%
$\longrightarrow$    
    \begin{minipage}{0.3\textwidth}
        \includegraphics[width=0.75\textwidth, height=0.18\textheight]{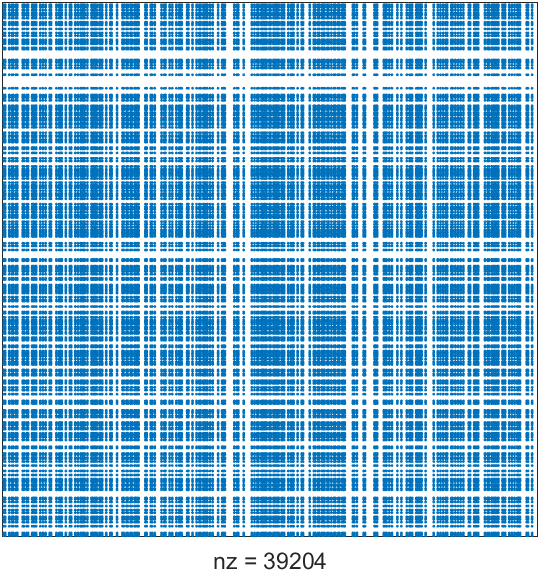}
    \end{minipage}%
    \caption{{\bf Left} A dense $I \times J$ matrix $\boldsymbol{\phi}$ of random values. {\bf Center} A block-matrix sparse $\boldsymbol{\phi} $ with $\frac{1}{4}IJ$ random entries. {\bf Right} The sparse version of $\boldsymbol{\phi}$  with the rows and columns randomly permuted to distribute the non-zero entries throughout the full image matrix.}
    \label{fig:dense_sparse_phi}
\end{figure}
However, using a sparse-$\boldsymbol{\phi}_{0}$ set for noisy or complicated images can result in unwanted boundary gaps in the final edge set. For many images, a useful approach is to scatter $IJ$ random points for the first RPI run, which in theory should find the general locations of object boundaries. The subsequent RPI runs can then be re-initialized with sparse random sets, which can be though of as filling in gaps along the boundary without adding too many extra points to regions where objects are close together.

\subsection{Post-processing example}
\label{ss:mpri_results}

\begin{figure}[!h]
    \centering
    \includegraphics[width=0.3\textwidth]{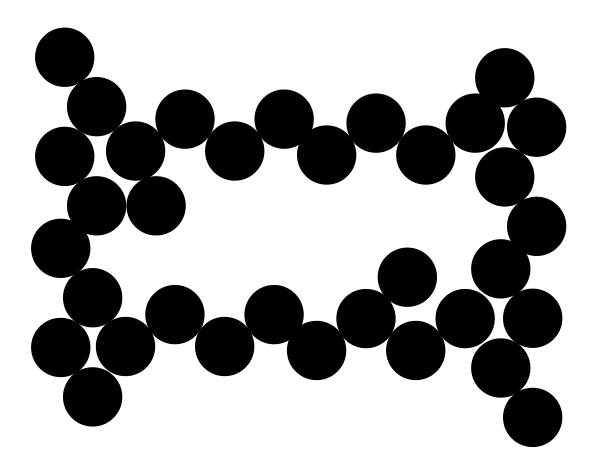}
    \caption[Sample synthetic image]{The original synthetic image}
    \label{fig:circles_synth}
\end{figure}

We demonstrate the efficacy of post-processing by segmenting the synthetic image shown in Figure \ref{fig:circles_synth} using the $m$-RPI method.

We initialize the first run of the RPI method using a dense random point set and re-initialize subsequent RPI runs with sparse random sets where $\alpha=0.25$ is the proportion of non-zero elements. Both the dense and sparse random sets have entries distributed normally as $\phi_{ij} \sim N(0,\sigma^{2})$ where $\sigma = 0.01$. We restart the method every $k=8$ iterations and iterate until a total of 15 RPI runs are completed. 
Note that each post-processing step is applied sequentially to the segmentation matrix in the order presented. 

\begin{figure}[!h]
    \centering
    \includegraphics[width=0.3\textwidth]{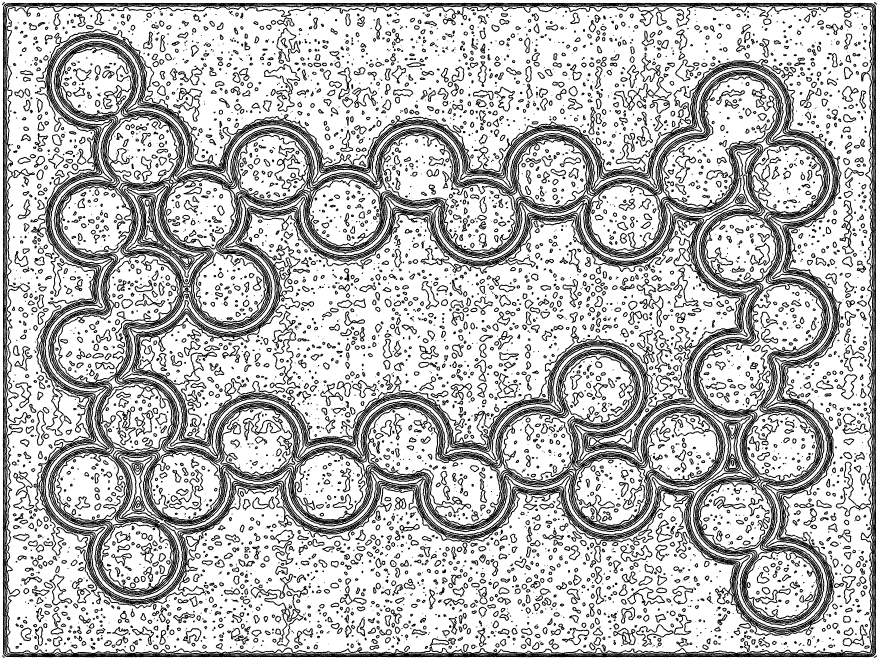}
    \caption{Segmentation for Figure \ref{fig:circles_synth} after averaging $15$ runs of the $m$-RPI method.}
    \label{fig:seg1_meanphi}
\end{figure}

We start by averaging $15$ $m$-RPI runs to find the average segmentation matrix $\overline{\boldsymbol{\phi}}$ as described in Section \ref{s:intro_rpi}. The resulting segmentation in Figure \ref{fig:seg1_meanphi} clearly outlines the boundary of object in the image, but this segmentation matrix also contains excessive noise.
Normalizing the entries of $\overline{\boldsymbol{\phi}}$ as in \eqref{e:norm_phi} pushes all of the values to the range $0 \leq \big \vert \frac{\boldsymbol{\overline{\phi}}}{||\boldsymbol{\overline{\phi}}||}  \big \vert \leq 1$. This does not de-noise the image but it accentuates the boundary of the object we wish to find.
\begin{figure}[!h]
    \centering
    \includegraphics[width=0.3\textwidth]{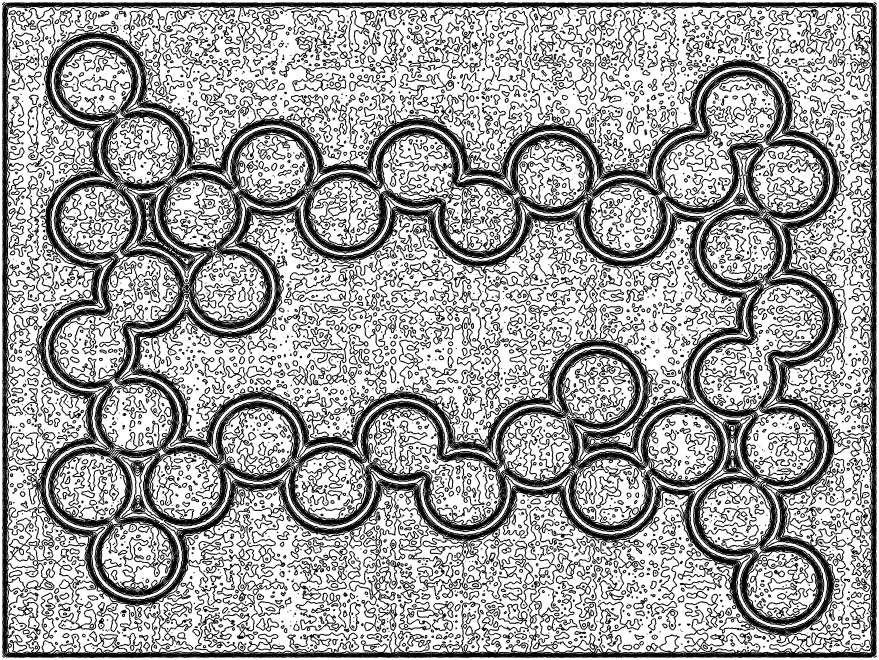}
    \caption{Normalization of $\overline{\boldsymbol{\phi}}$.}
    \label{fig:seg2_normphi}
\end{figure}
We then threshold to approximate the zero-level set $\overline{\boldsymbol{\phi}}_{\text{\tiny log}}$ as in \eqref{e:phi_log} by taking a small range of values around zero. In this example we use $[p_{\ell},p_{u}] = [-0.175,0.075]$. Experimental evidence suggests that using a larger proportion of negative entries (values within the object in question) gives a cleaner segmentation.
\begin{figure}[!h]
    \centering
    \includegraphics[width=0.3\textwidth]{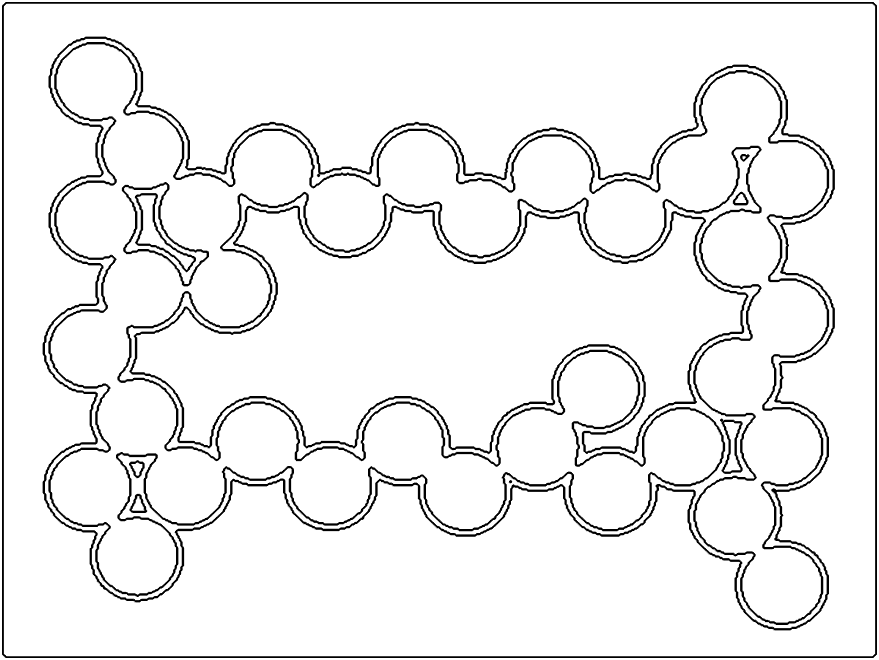}
    \caption{Narrow band approximation to the zero-level set.}
    \label{fig:seg3_philog}
\end{figure}
The thresholding step effectively de-noises the edge set but also results in a \emph{doubled} boundary curve around the object. We apply both curve smoothing and thinning to fill and thin the boundary, producing the final image segmentation for the sample image. (Note that the built-in Matlab\textsuperscript{\tiny \textregistered} function \texttt{bwmorph} was used to implement the curve smoothing and thinning routines. The morphological operation \texttt{majority} with the additional option \texttt{Inf} was used to smooth the curve while \texttt{thin}, which implements the thinning algorithm described in Section \ref{s:major_thin}, was used to skeletonize the boundary.)
\begin{figure}[!t]
    \centering
    \hfill
    \begin{minipage}{0.4\textwidth}
    \includegraphics[width = 0.75 \textwidth, height = 0.15 \textheight]{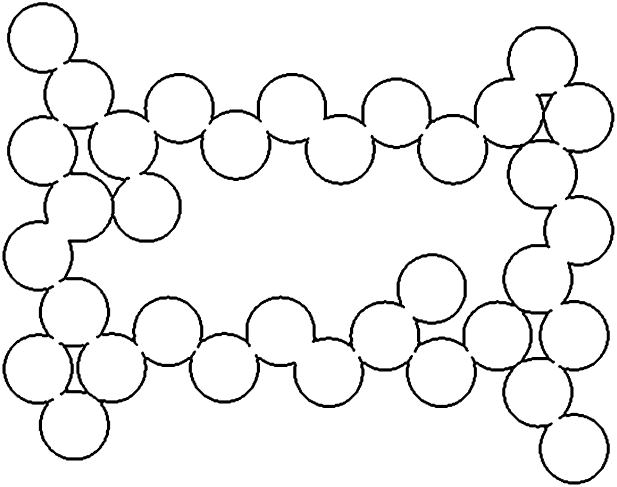}    
    \end{minipage}    
    \hfill
    \begin{minipage}{0.4\textwidth}
    \includegraphics[width = 0.75\textwidth, height = 0.15 \textheight]{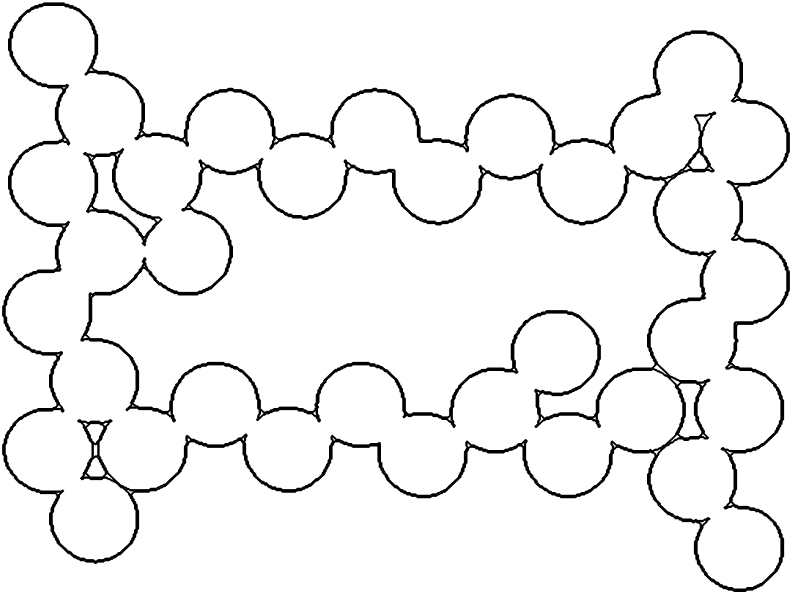}    
    \end{minipage}
    \caption[Comparison of ground truth segmentation for Figure \ref{fig:circles_synth} to $m$-RPI segmentation]{\emph{Left:} The ground truth segmentation of the image in Figure \ref{fig:circles_synth}. \emph{Right:} The final segmentation obtained with the $m$-RPI method after the boundary curve has been smoothed and thinned.}
    \label{fig:seg4_finalphi}
\end{figure}

The expected segmentation of the image (the ``ground truth") is included in Figure \ref{fig:seg4_finalphi} for comparison.

\section{Experimental Results: Comparison to the Canny Method}

Returning to the image of the spider from Section \ref{s:intro_rpi} and now using the $m$-RPI method, we successfully segmented the image using $15$-RPI runs. In addition to finding the full boundary of the spider, including the troublesome region around his legs, the $m$-RPI method located additional details within the image -- such as the pattern of spots on the spider's back and texture of the leaf he is standing on -- that were previously missed by the classical level set method.
\begin{figure}[h!]
    \centering
    \includegraphics[width=\textwidth]{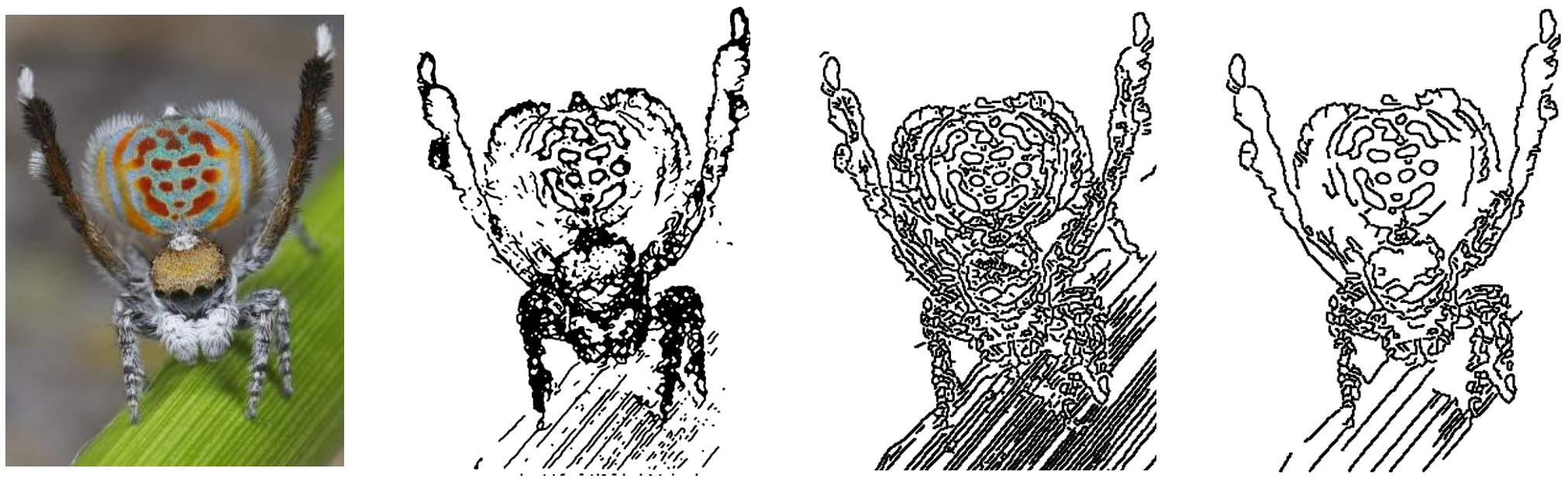}
    \caption{Comparison of the $m$-RPI method (center-left) to the Canny method when the Canny parameters are both default (center-right) and optimized (far-right).}
\end{figure}

As image segmentation is an ill posed problem, each segmentation task depends heavily on the image that will be segmented (``synthetic" or ``real") and the type of information sought from the image. In general, it is difficult to establish any consistent criteria for evaluating segmentation algorithms. While synthetic images often allow for pixel-to-pixel comparisons of a ground truth to the segmented image, these are not representative of most real images of interest and are not useful for evaluating an algorithm on real datasets. 

To evaluate the performance of the $m$-RPI method on the spider image, we make a visual comparison to the segmentation results found using the Canny Edge Detector, which is considered state-of-the-art. Notice that two Canny results are given. In an effort to compare the methods as honestly as possible, significant time and effort was dedicated to optimizing the parameters used in the Canny Edge Detector provided by Matlab.\footnote{We chose an upper thresholding value of $0.2$ and a lower thresholding value of $0.1$ as optimal values.} It should be noted that, prior to optimizing the Canny thresholding values in Matlab, the $m$-RPI method equalled or surpassed the Canny segmentation of most images that we examined. When the $m$-RPI results for the spider are compared to the results of the optimized Canny method, we see similar levels of detail in the segmentation. However, the $m$-RPI method clearly outperforms the Canny segmentation obtained with the default parameter settings.

We conclude by demonstrating the results of segmenting real images with the $m$-RPI method and we visually compare our results to the Canny Edge Detector.\footnote{Test images belong to the authors and may not be re-used without permission.}


\section{Conclusion}
\label{s:concl}

The Canny Method is considered the state-of-the-art image segmentation algorithm, but the performance of this method depends heavily on the chosen upper and lower thresholding values that are used. Careful selection of the parameters used in the Canny algorithm is required for the method to consistently perform well, and therefore significant experimentation is required to optimize the thresholding values. Because of this, our algorithm has the advantage of successfully segmenting many real image without extensive parameter tuning and may be easily implemented for many image segmentation tasks. 

While the Canny Method in Matlab and Python segments images more quickly (clock time) than our current $m$-RPI code does -- for example, the spider image was segmented with Matlab's built-in Canny in approximately 6 seconds while the $m$-RPI method with post-processing took approximately 29 second -- the $m$-RPI method promises substantial speed-up under a parallelization of the code as no RPI run depends on any of the others.

\newpage 
\setcounter{section}{4}
\subsection{Comparison to Canny Segmentation}
\label{ss:gallery}


\begin{figure}[h!]
    \centering
    \includegraphics[width=\textwidth]{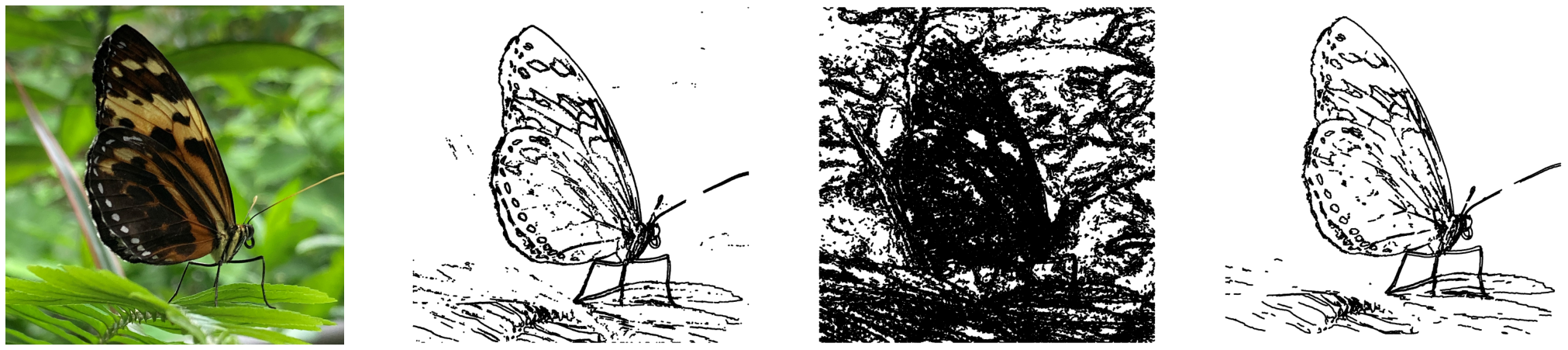}
\end{figure}

\begin{figure}[h!]
    \centering
    \includegraphics[width=\textwidth]{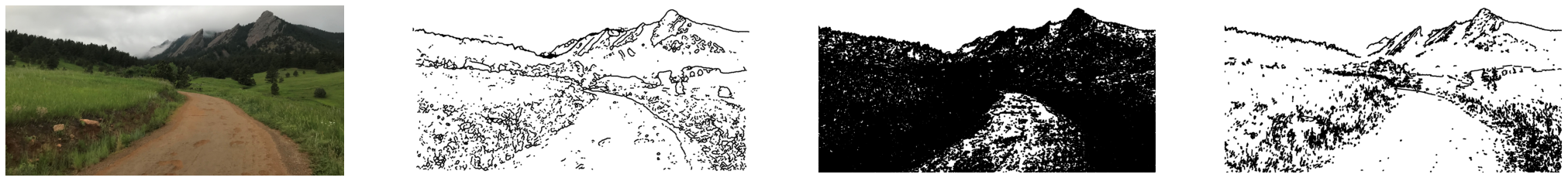}
\end{figure}

\begin{figure}[h!]
    \centering
    \includegraphics[width=\textwidth]{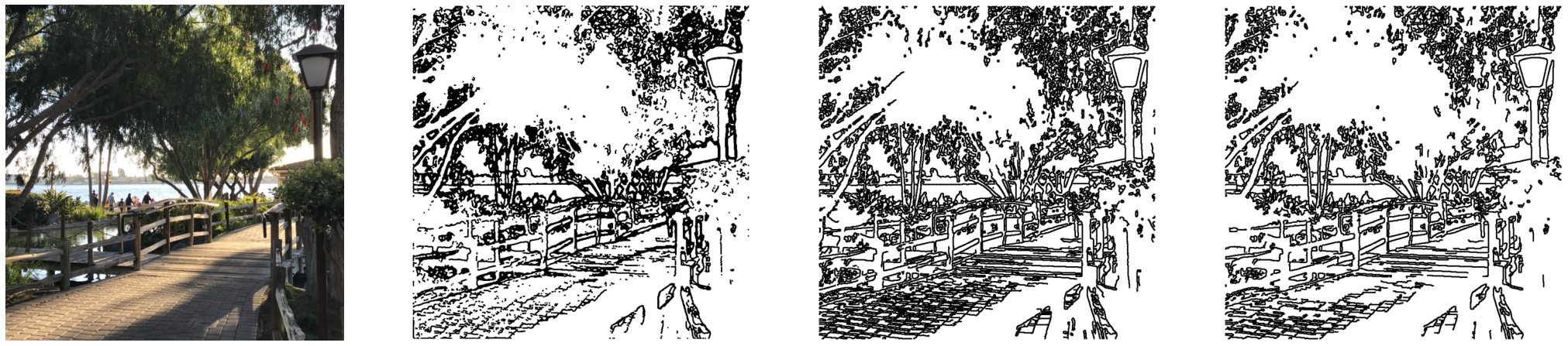}
\end{figure}

\begin{figure}[h!]
    \centering
    \includegraphics[width=\textwidth]{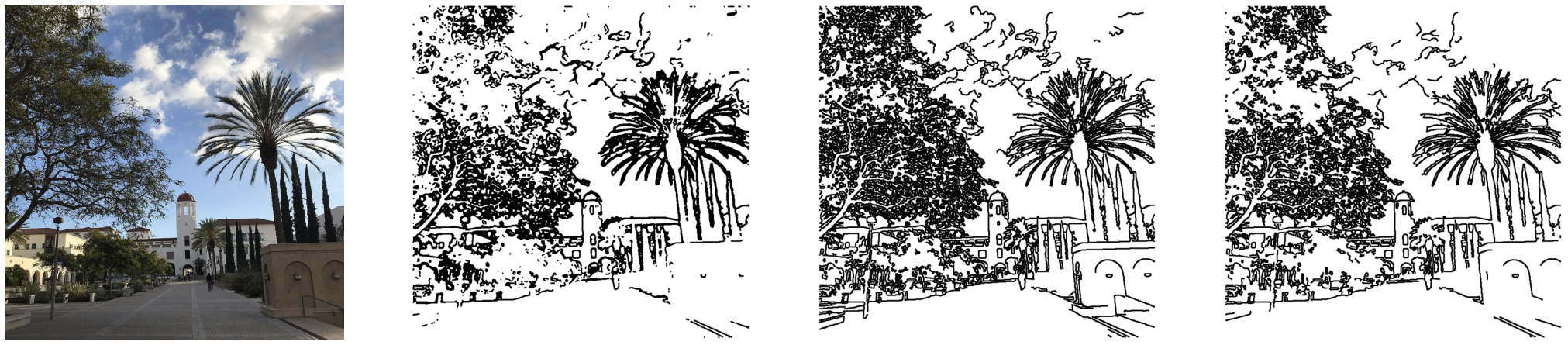}
\end{figure}

\begin{figure}[h!]
    \centering
    \includegraphics[width=\textwidth]{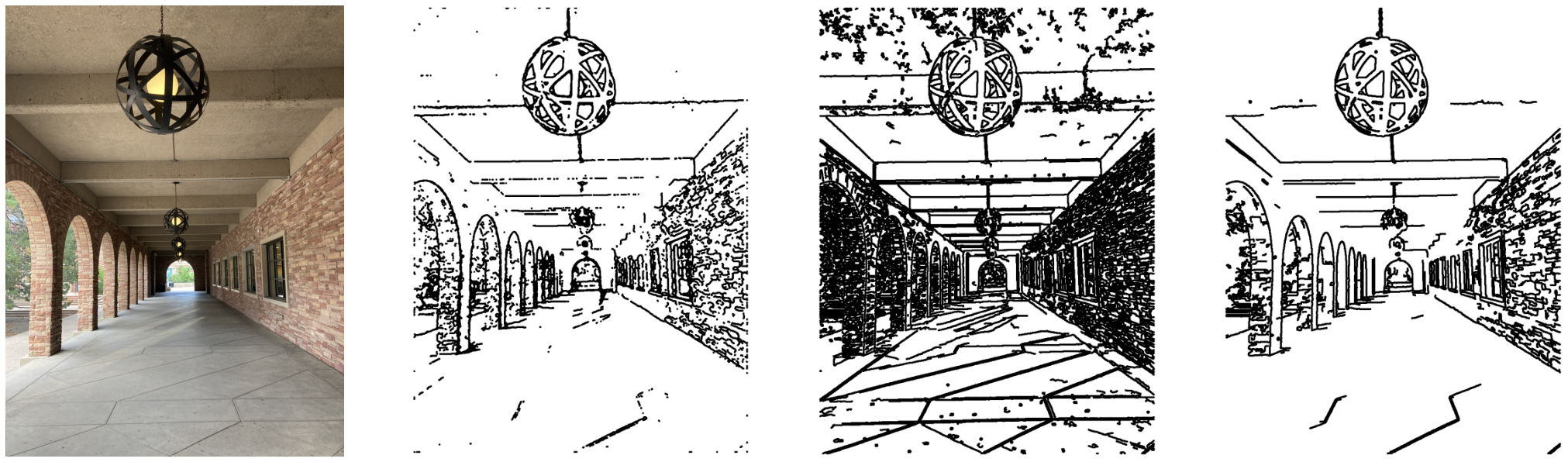}
\end{figure}

\newpage 
\begin{figure}[h!]
    \centering
    \includegraphics[width=\textwidth, height=0.175\textheight]{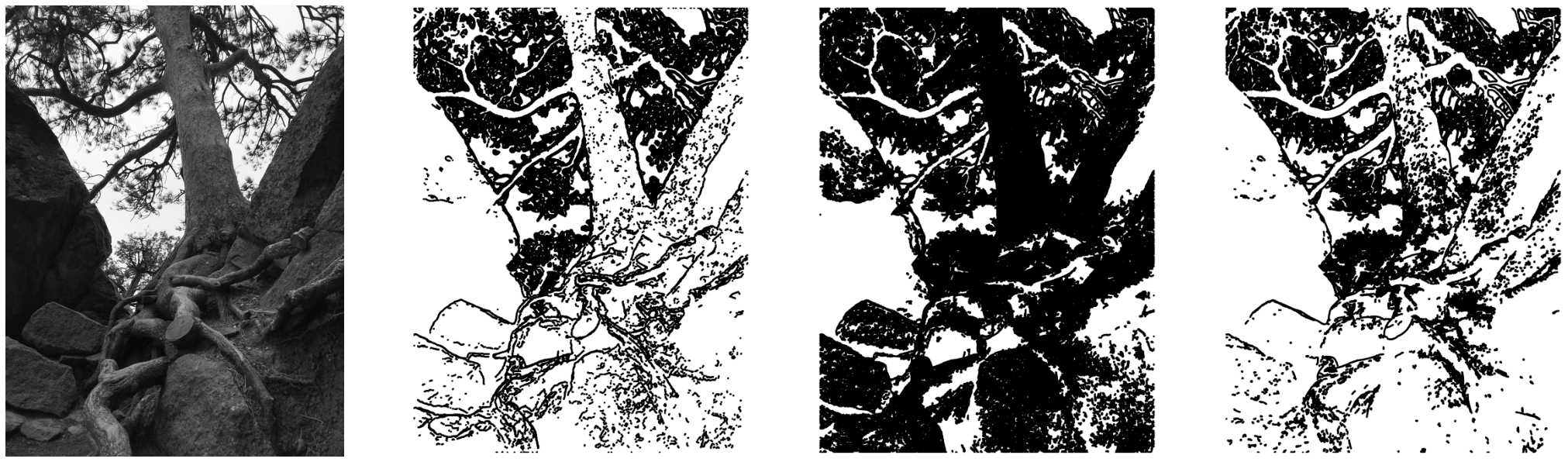}
\end{figure}

\begin{figure}[h!]
    \centering
    \includegraphics[width=\textwidth]{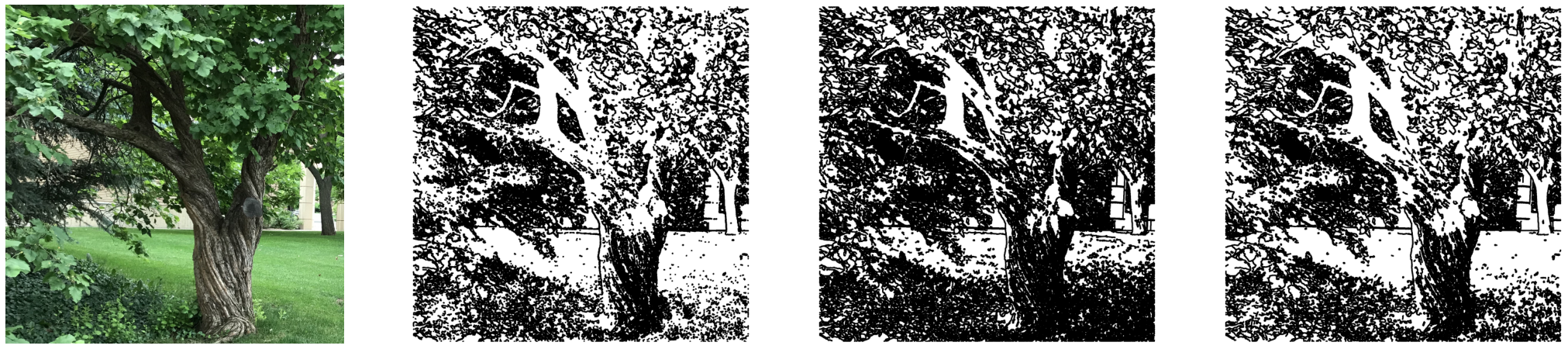}
\end{figure}

\begin{figure}[h!]
    \centering
    \includegraphics[width=\textwidth]{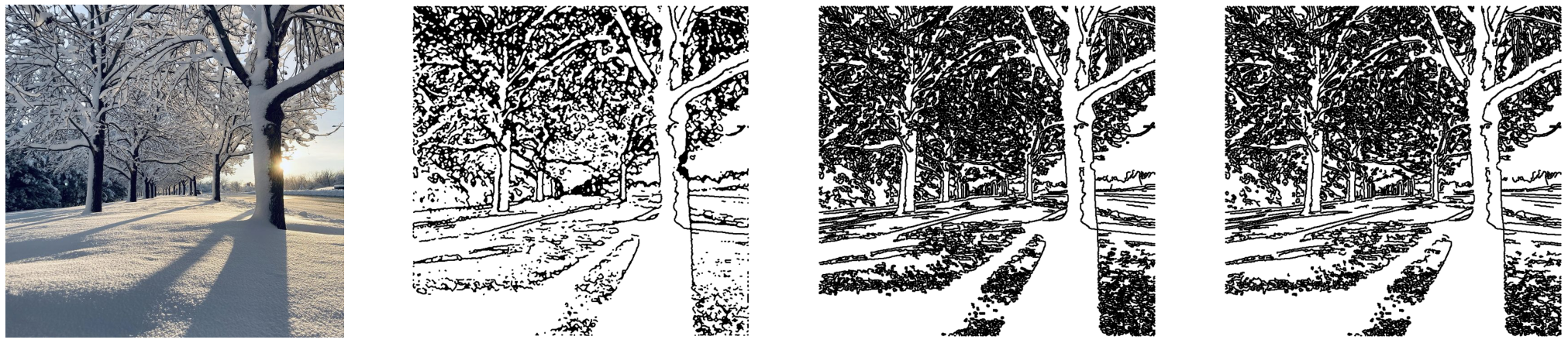}
\end{figure}

\begin{figure}[h!]
    \centering
    \includegraphics[width=\textwidth]{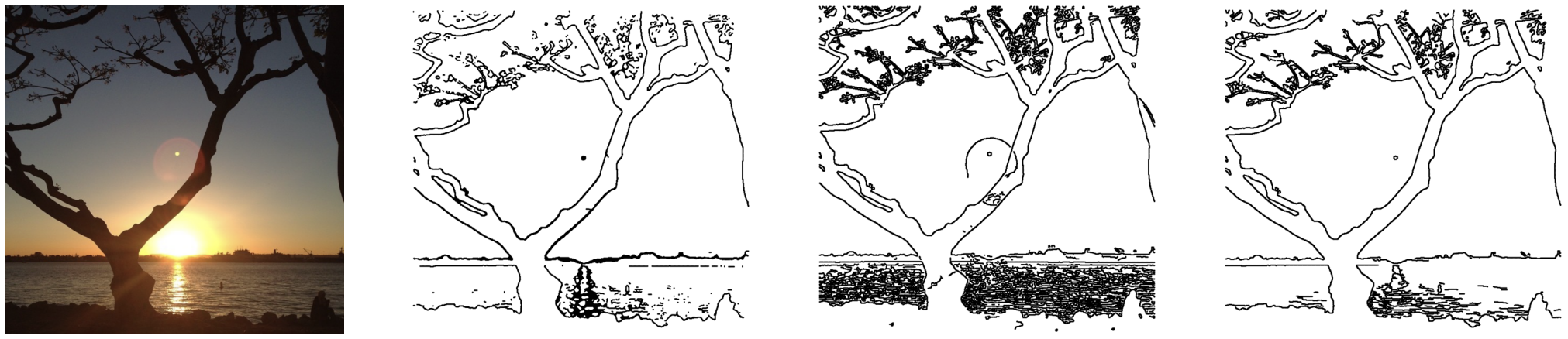}
\end{figure}

\begin{figure}[h!]
    \centering
    \includegraphics[width=\textwidth]{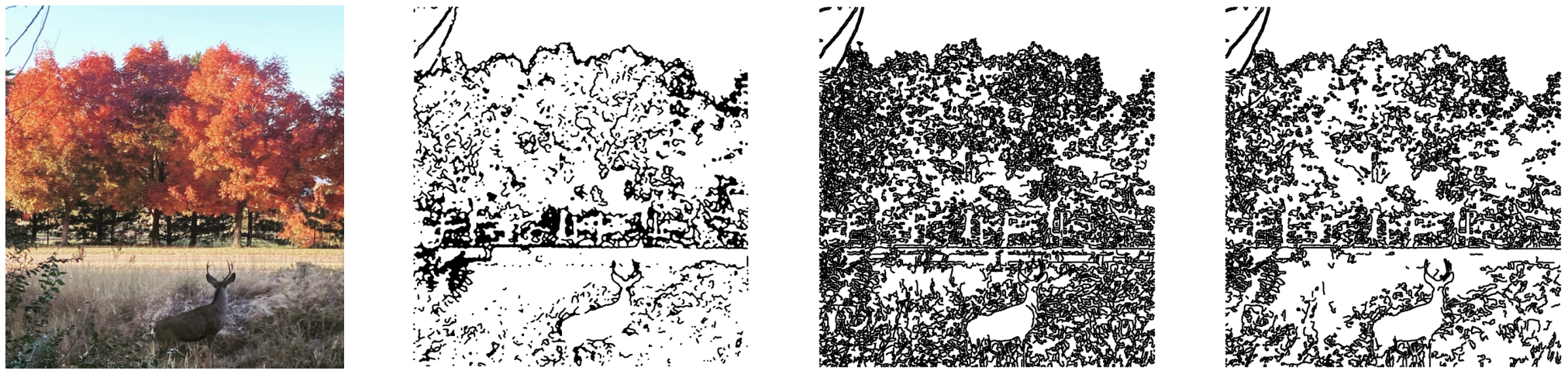}
\end{figure}

\newpage 

\bibliographystyle{vancouver}
\bibliography{masterbib.bib}

\end{document}

%% file: macros.tex
\newsavebox{\junk}
\savebox{\junk}[1.6mm]{\hbox{$|\!|\!|$}}

\def\bbbz{{\mathchoice {\hbox{$\sf\textstyle Z\kern-0.4em Z$}}
{\hbox{$\sf\textstyle Z\kern-0.4em Z$}}
{\hbox{$\sf\scriptstyle Z\kern-0.3em Z$}}
{\hbox{$\sf\scriptscriptstyle Z\kern-0.2em Z$}}}}
\def\sq{\hbox{\rlap{$\sqcap$}$\sqcup$}}

\usepackage{amsmath} 
\usepackage{amsfonts}
\usepackage{latexsym}
\usepackage{bm}

\newcommand{\ben}{\begin{enumerate}}
\newcommand{\een}{\end{enumerate}}
\newcommand{\bit}{\begin{itemize}}
\newcommand{\eit}{\end{itemize}}

\newtheorem{theorem}{Theorem}[section]
\newtheorem{proposition}[theorem]{Proposition}

\newcommand{\ba}{\begin{array}{rcl}}
\newcommand{\ea}{\end{array}}
\newcommand{\bt}{\begin{theorem}}
\newcommand{\et}{\end{theorem}}
\newcommand{\bd}{\begin{description}}
\newcommand{\ed}{\end{description}}

\def\eq#1/{(\ref{e:#1})}

\def\beq{\begin{equation}}
\def\eeq{\end{equation}}
\def\beqa{\begin{eqnarray}}
\def\eeqa{\end{eqnarray}}

\def\qed{\ifmmode\sq\else{\unskip\nobreak\hfil
\penalty50\hskip1em\null\nobreak\hfil\sq
\parfillskip=0pt\finalhyphendemerits=0\endgraf}\fi}

\def\sqr#1#2{{\vcenter{\hrule height.#2pt
      \hbox{\vrule width.#2pt height#1pt \kern#1pt
         \vrule width.#2pt}
       \hrule height.#2pt}}}

\newcommand{\bc}{\begin{corollary}}
\newcommand{\ec}{\end{corollary}}

\newcommand{\bp}{\begin{proposition}}
\newcommand{\ep}{\end{proposition}}

\def\eye(#1){{\bf (#1)}\quad}

\def\taboo#1{{{}_{#1}}}

\def\0P{\taboo{0}P}
\def\0Pn{\taboo{0}P^n}

\usepackage{bbm}
\usepackage{amsmath}
\DeclareSymbolFont{largesymbols}{OMX}{yhex}{m}{n}
\DeclareMathAccent{\widewidehat}{\mathord}{largesymbols}{"62}

\usepackage{color}
\usepackage{graphicx}
\usepackage[position=t,singlelinecheck=off]{subfig}
\usepackage{caption}
\usepackage{tikz}
\usepackage[version=4]{mhchem}
\usepackage{amsmath,amssymb,tikz-cd}
\usepackage[titletoc]{appendix}

\usepackage{comment}

\DeclareMathAlphabet{\mathpzc}{OT1}{pzc}{m}{it}

\date{}

\usepackage{mathtools}
\usepackage{extarrows}
\usetikzlibrary{matrix}
\usepackage{multicol}

\usepackage{tikz}
\usepackage{xcolor}
\usetikzlibrary{arrows}
\usetikzlibrary{positioning}
\usetikzlibrary{matrix,fit}
\pgfkeys{tikz/mymatrix/.style={matrix of math nodes,inner sep=0pt,row sep=0em,column sep=0em,nodes={inner sep=6pt}}}

\usepackage{dutchcal}

\usepackage{tikz}
\usepackage{xcolor}
\usetikzlibrary{arrows}
\usetikzlibrary{decorations.markings}
\usetikzlibrary{positioning}
\usetikzlibrary{matrix,fit}
\pgfkeys{tikz/mymatrix/.style={matrix of math nodes,inner sep=0pt,row sep=0em,column sep=0em,nodes={inner sep=6pt}}}

\usepackage{algorithm}
\usepackage[noend]{algpseudocode}

\usepackage{colortbl}

\usepackage[format=plain,
            labelfont={bf,it},
            textfont=it,
            margin=0.05in,
            font=small]{caption}

\newcommand{\ds}{\displaystyle}

